\documentclass[10pt,twocolumn,letterpaper]{article}

\usepackage{iccv}
\usepackage{times}
\usepackage{epsfig}
\usepackage{graphicx}
\usepackage{amsmath}
\usepackage{amssymb}
\newcommand{\shortmethodname}{DONNAv2 }
\usepackage{float}
\usepackage{algorithm}
\usepackage{algpseudocode}
\usepackage{adjustbox}
\usepackage{subfig}
\usepackage{caption}
\usepackage{subcaption}
\usepackage{amssymb}

\usepackage[pagebackref=true,breaklinks=true,letterpaper=true,colorlinks,bookmarks=false]{hyperref}
\iccvfinalcopy 

\def\iccvPaperID{41} 

\ificcvfinal\fi

\begin{document}

\title{\shortmethodname - Lightweight Neural Architecture Search for Vision tasks}

\author{Sweta Priyadarshi, Tianyu Jiang, Hsin-Pai Cheng, Sendil Krishna, Viswanath Ganapathy, Chirag Patel\\
Qualcomm AI Research\thanks{ Qualcomm AI Research is an initiative of Qualcomm Technologies,
Inc.}\\
San Diego, CA, USA 92121\\
{\tt\small \{swetpriy, tianyuj, hsinpaic, sendilk, viswgana, cpatel\}@qti.qualcomm.com}
}

\maketitle
\ificcvfinal\thispagestyle{empty}\fi

\begin{abstract}
   With the growing demand for vision applications and deployment across edge devices, the development of hardware-friendly architectures that maintain performance during device deployment becomes crucial.
   Neural architecture search (NAS) techniques explore various approaches to discover efficient architectures for diverse learning tasks in a computationally efficient manner. In this paper, we present the next-generation neural architecture design for computationally efficient neural architecture distillation - \shortmethodname. Conventional NAS algorithms rely on a computationally extensive stage where an accuracy predictor is learned to estimate model performance within search space. This building of accuracy predictors helps them predict the performance of models that are not being finetuned.  Here, we have developed an elegant approach to eliminate building the accuracy predictor and extend DONNA to a computationally efficient setting. The loss metric of individual blocks forming the network serves as the surrogate performance measure for the sampled models in the NAS search stage. To validate the performance of \shortmethodname we have performed extensive experiments involving a range of diverse vision tasks including classification, object detection, image denoising, super-resolution, and panoptic perception network (YOLOP). The hardware-in-the-loop experiments were carried out using the Samsung Galaxy S10 mobile platform. Notably, \shortmethodname reduces the computational cost of DONNA by 10x for the larger datasets. Furthermore, to improve the quality of NAS search space, \shortmethodname leverages a block knowledge distillation filter to remove blocks with high inference costs. 
\end{abstract}

\section{Introduction}

\label{sec:intro}
Computer vision algorithms are being widely deployed on edge devices for several real-world applications including medicine, XR-VR technology, visual perception, and autonomous driving. However, computer vision algorithms based on deep learning require significant computational resources. Therefore, efficient search for deep learning architecture has attracted a lot of attention. Most of these NAS efforts are agnostic to the requirements of resource-constrained edge devices. Further, current NAS methods that operate over large search spaces are computationally very expensive to generate the optimized models. NAS based on block knowledge distillation (BKD) \cite{blakeney2020parallel, donna, layernas} scales well over large search spaces in a computationally efficient manner. In this work, we leverage BKD for hardware-aware NAS. The core process of our NAS approach based on BKD consists of building replacement blocks, building accuracy predictors, predicting the accuracy of models, and based on their cost(flops, parameters, latency on hardware), the models are picked based on the trade-off between predicted accuracy and cost.  In multiple studies, it appears that the stage of building the accuracy predictor is the most expensive bottleneck of the pipeline. Researchers have worked on making the accuracy predictor stage efficient by utilizing regression or ranking methods. Nevertheless, the majority of the computation time for the NAS pipeline is still taken up by the accuracy predictor stage. Our work \shortmethodname aims at reducing the search space by identifying the redundant blocks and eliminating them from the search space. We have defined this method as the Blockwise Knowledge Distillation filtering stage. Furthermore, we aimed at removing the accuracy predictor stage, which was by far the most computationally expensive component of the DONNA pipeline. Our work \shortmethodname brings a more sophisticated method of approximating blockwise losses to network losses.\\
Many NAS studies have focused on optimizing models for hardware-agnostic complexity metrics like flops (MACs). But some of the analyses \cite{analysis}, indicate that flops do not always translate linearly to the latency or the power of the model. To find the best architecture for a given use case, and a given hardware, it is important to specifically optimize models to minimize latency and energy consumption for on-device performance. Many NAS performs with a lookup table that is reporting per-layer latency and is approximated to full model latency. Here, the assumption is that the linear sum of latency would be model latency which does not hold true always. We have hardware in the loop to optimize models for a given hardware. But unlike many expensive methods, \shortmethodname tends to provide optimal neural networks at a lower complexity for a similar diverse search space. In our work, we have compared the time complexity saved by pivoting to the approximation method using mean square error (MSE) loss rather than training an accuracy predictor model.\\
We have described our paper through the \shortmethodname pipeline that comprises of - Block knowledge Distillation (BKD), BKD Filtering, Evolutionary Search, and Finetuning for a Galaxy S10 mobile platform. Finally, we extend our paper to cover five vision tasks to show how \shortmethodname led to an optimal compressed model without losing accuracy. The vision tasks highlighting the benefits of \shortmethodname are but not limited to image classification, object detection, super-resolution, image denoising, and multitask network.

\begin{figure*}[t][h]
   \centering
   \includegraphics[width=1\textwidth]{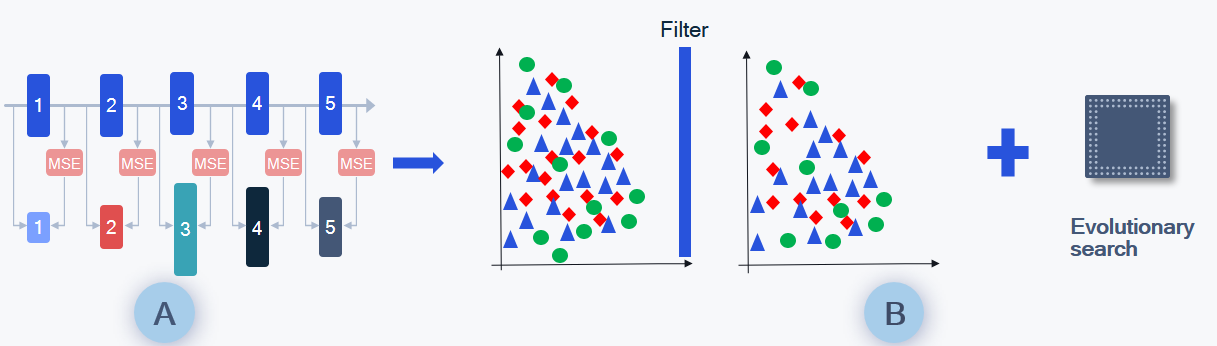}
   \caption{\shortmethodname Pipeline - includes Stage A which is composed of search space definition and Block Knowledge Distillation(BKD), and Stage B which includes BKD Filtering, Evolutionary search(hardware in the loop), and Finetuning.
}
   \label{fig:pipeline}
\end{figure*}

\section{Related Work}

We can delve into the historical progression of NAS to trace its evolution from initially computationally expensive methods involving diverse search space \cite{mnasnet, rl, learnable} to low computation methods with very small search space \cite{ofa, single_path_nas}. DONNA \cite{donna}, explored approaches to reduce the computational burden using a block-based search space. Recent study \cite{chau2022blox} has also validated the efficacy of NAS approach developed in DONNA. 
Here, \shortmethodname, we aim to further reduce the computation time of the search while keeping the search space similar to DONNA. Mobile neural architecture search (MNAS) \cite{mnasnet} is an expensive method that 
requires around 40,000 epochs to perform a single search.
Other attempts for NAS included differentiable architecture search methods such as DARTS \cite{darts}, FBNet \cite{fbnet}, FBNetV2 \cite{fbnetv2}, ProxylessNAS \cite{proxylessnas}, AtomNAS \cite{atomnas} and 
Single-Path NAS\cite{single_path_nas} that simultaneously optimize the weights of a large super-net and its architectural parameters. However, in these cases the granularity of the search level suffers and methods need to be repeated in every scenario or when the search space changes. There have been studies \cite{zerocost, one_shot, hardwareaware, layernas} to construct proxies for ranking models using the search space. These include attempts based on zero-shot proxy \cite{zerocost} and one-shot proxy NAS \cite{one_shot}. A similar approach LANA \cite{hardwareaware}, also leverages the loss function as a proxy method to rank the model. A recent work \cite{layernas} explores a hardware-aware search by translating the multi-objective optimization problem into a combinatorial optimization problem. However, this approach assumes chain-structured NAS and is not readily applicable to more general architectures. Our \shortmethodname builds on the idea of using the loss function as the proxy with the following enhancements:
\begin{itemize}
        \item enables hardware-aware search in a diverse search space with the hardware in the loop for latency measurements. Earlier studies leveraged a linear sum of the pre-computed feature layer latencies to estimate the latency of a deep learning model. However, this does not capture the true latency when compilers leveraged a depth-first search. 
        \item \shortmethodname is scalable when the search is expanded or the hardware platform changes. 
        \item \shortmethodname converges 9x faster during the finetuning stage while achieving a similar accuracy compared to training-from-scratch (\cite{dna}).

\end{itemize}

\section{\shortmethodname - Lightweight NAS}
\shortmethodname follows the steps in DONNA, while eliminating the accuracy predictor stage and introducing a blockwise knowledge distillation (BKD) filtering stage. In \shortmethodname, we start by defining a search space and then building a BKD library which gets further filtered out by a BKD filter. These filtered blocks from the BKD library are utilized by the evolutionary search phase to find the Pareto-optimal network architectures for any specific scenario using the loss metric of individual blocks. Finally, the predicted Pareto-optimal architectures are fine-tuned to full accuracy for deployment.

\subsection{Search Space}
Search Space in \shortmethodname follows a block-level architecture and only parameters within the blocks are varied. A collection of blocks to build candidate networks are generated based on user-defined blocks and 
the associated parameters.  To determine a  suitable search space, we include diverse macro-architectural network parameters such as layer types, attention mechanisms, and channel widths. Furthermore,  micro-architectural parameters such as cell repeats within a block, kernel sizes of layers within cells, and in-cell expansion rates were also utilized. In our experiments, the cardinality of the search space was of the order of 1e14. The larger the search space, the higher the chances of identifying hardware-friendly and 
performance-achieving networks.

\subsection{Blockwise Knowledge Distillation}
Blockwise Knowledge Distillation (BKD) is the first building block of the lightweight NAS, \shortmethodname. Unlike DONNA, \shortmethodname uses BKD not for building an accuracy predictor, but as an input to produce a surrogate metric to generate the Pareto optimal curve. The BKD stage generates a Block Library with pre-trained weights and loss metrics for each of the option blocks $B_{n,m}$ that is used as the replacement. To build the BKD library, each block option $B_{n,m}$ is replaced in the blocks of the mothernet and trained as a student model using the mothernet block $B_n$ as a teacher. The MSE loss between the teacher’s output feature map $Y_{n}$ and the student’s output feature map $\bar{Y}_{n, m}$ is used as the surrogate metric in the evolutionary search stage and the BKD filtering stage. One epoch of complete dataset training is employed at this stage for building the BKD library and is denoted as 1e. The pre-trained weights at this stage help in faster convergence while finetuning the model.

\subsection{Blockwise Knowledge Distillation Filtering}
\begin{figure}[t][h]
  \centering
   \includegraphics[width=0.9\linewidth]{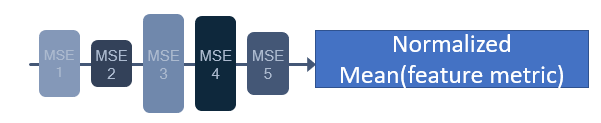}
   \caption{Metric developed using the MSE loss of the blocks forming the network, for the Pareto optimal curve}
   \label{fig:bkdfil}
\end{figure}

Blockwise Knowledge Distillation Filtering method aims to identify and drop 
the inefficient blocks based on the optimization strategy. Here, the optimization strategy is defined as the cost of the optimized model in terms of flops, latency, power consumption, etc. The BKD filtering stage retains only blocks with a minimum cost ratio with respect to the associated blocks in the reference model. Blocks with minimum cost ratio will retain performance-achieving efficient candidate models during the evolutionary search. The cost ratio of a block is estimated for a given loss metric. Retaining only blocks with the best cost ratio reduces the number of blocks and thereby the cardinality of the search space. 
However, it is important to note that block filtering does not eliminate good models in the sample space. We have validated this with experiments across several learning tasks. In Figure \ref{fig:bkdfil}, the legend id tells the blocks of a particular layer we are filtering and the blue dots represent the blocks that are retained and grey dots represent the blocks that would be dropped.

In Algorithm 1, we have described the steps in detail.

\begin{algorithm}[h]
\caption{BKD filtering}\label{alg:cap}
\textbf{Input:}
BKD library, threshold = D.\\
\hspace*{\algorithmicindent}\hspace*{\algorithmicindent} $B_{(n,m)}$ is the $m^t$$^h$ potential replacement out of M\\
\hspace*{\algorithmicindent}\hspace*{\algorithmicindent}choices for block $B_n$ in the mothernet model.
\begin{algorithmic}
\For i = 1 to m do
\State  $L_{(n,m)}$ = Calculate the block $B_{(n,m)}$ inference cost of model 
\State $MSE_{(n,m)}$ = Calculate the block $B_{(n,m)}$ MSE loss of model w.r.t mothernet
\State $C_{(n,m)}$ = Calculate the ratio of $L_{(n,m)}$ w.r.t mothernet block inference cost 
\State Plot cost ratio vs MSE on a plot
\State Discard the blocks at each MSE loss with higher inference cost based on the threshold D.
\State \textbf{OUTPUT:} Obtain new BKD filtered library 
\EndFor
\end{algorithmic}
\end{algorithm}

\begin{figure}[t]
  \centering
   \includegraphics[width=0.9\linewidth]{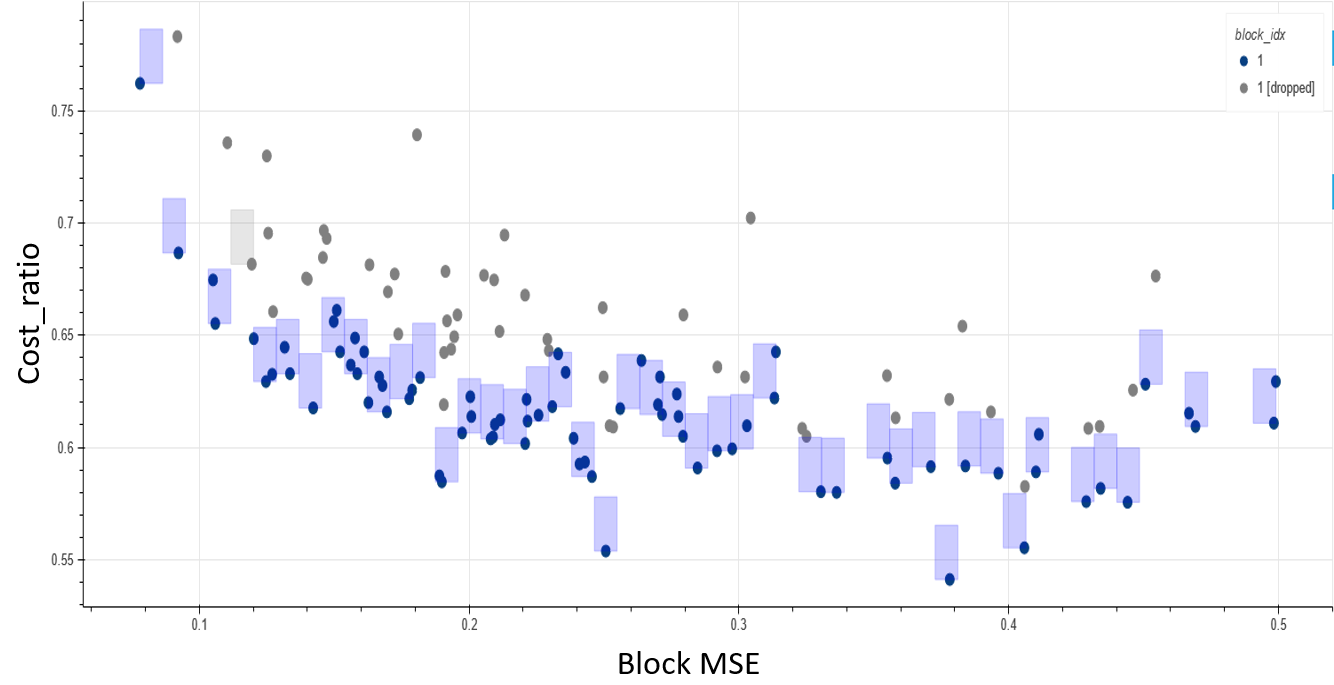}
   \caption{BKD Filtering - The x-axis is the surrogate loss metric and the y-axis is the cost ratio(flops/latency on device) between the replacement blocks and the mothernet
}
   \label{fig:bkdfilter}
\end{figure}

\subsection{Evolutionary Search}

Evolutionary search utilizes the MSE loss metric as a surrogate measure. Here, in contrast to DONNA, we lack
the predicted accuracy of the candidate models from the Pareto front. The performance  
of the model is approximated as the sum of the MSE of the blocks constituting the model. We built the Pareto optimal curve by using this surrogate measure of the model.
Given the MSE loss metric of blocks from the block library and latency of the models formed by using the block options, the NSGA-II evolutionary algorithm is leveraged to find Pareto-optimal architectures that
minimize the model loss and cost. The cost utilized could be scenario-agnostic measures such as the number of operations (MAC) or the number of parameters in the network (params). The scenario-aware cost includes on-device latency, cycles of operation, and energy. In our experiments,  we have utilized on-device latency as a
cost function by using direct hardware measurements in the optimization loop. After obtaining the Pareto-optimal models, we selected the model with appropriate latency and finetuned the candidate model to obtain the final model.

\subsection{Finetuning}
Empirically it has been observed that the final architectures from the Pareto front curve converge faster than training from scratch when pre-trained with weights obtained from the BKD stage. It has been shown that EfficientNet-style models can converge in around 50 epochs as opposed to 450 epochs when trained from scratch.

\section{Experiments \&\ Results}
In this section we will discuss the detailed experimental evaluation of \shortmethodname:
across a set of diverse computer vision tasks.
The performance of \shortmethodname for Image classification, Object detection and super-resolution tasks were 
quantified with the Samsung Mobile hardware platform in the loop. The performance of \shortmethodname for image denoising and multitask network was validated using the number of operations(MAC). Importantly, all experiments demonstrated significant model compression with minimal performance degradation on an edge device. The performance of \shortmethodname is captured in terms of accuracy, on-device latency/MAC, and the number of epochs (defined as the sum of the number of epochs for training accuracy predictor, the number of epochs used for finetuning and building the block library (1e)). It is important to note that there is very few NAS methodology that has worked for diverse vision tasks. Many research focuses on NAS method for individual Vision tasks, but we aim to focus on key components that remain the same across the wide range of vision tasks and deliver the same method to be applied across various vision tasks. Details of each of the experiments are described below:

\subsection{Search Algorithm}

In this section, we have summarized the overall \shortmethodname setup in an algorithm format as shown in algorithm \ref{alg:cap}. It provides step by step setup details to perform the BKD based searching.

\begin{algorithm}[h]
\caption{\shortmethodname search}\label{alg:cap}
\textbf{Input:}
Begin with a baseline or \textbf{mothernet} network.\\
\hspace*{\algorithmicindent}\hspace*{\algorithmicindent}Split the baseline into stem, head and N blocks.\\
\hspace*{\algorithmicindent}\hspace*{\algorithmicindent} $B_(n,m)$ is the $m^t$$^h$ potential replacement out of M\\
\hspace*{\algorithmicindent}\hspace*{\algorithmicindent}choices for block $B_n$ in the baseline model.
\begin{algorithmic}
\For i = 1 to m do
\State \textbf{BKD:}Replace block $B_n$ with $B_(n,i)$ and train the \\
\hspace*{\algorithmicindent}\hspace*{\algorithmicindent}new architecture for 1 epoch of complete dataset.
\hspace*{\algorithmicindent}\hspace*{\algorithmicindent} Complete the step for all blocks and all options \\
\hspace*{\algorithmicindent}\hspace*{\algorithmicindent} and construct a BKD library with MSE loss of \\
\hspace*{\algorithmicindent}\hspace*{\algorithmicindent} replacement blocks w.r.t the mothernet. 
\EndFor
\State \textbf{BKD Filetring :}
Perform BKD filtering to remove redundant blocks as explained in Section \textbf{3.2}
\State \textbf{Evolutionary Search:}
\State \textbf{Input:}
Population Size= E, number of search steps = T\\
BKD Library 

\For i = 1 to T do
\State  Randomly sample E networks Ft from networks composed of $B_(n,m)$ blocks
\State Compute inference cost \& MSE of the sampled model Ft
\State Retain models with lowest MSE loss in each iteration at different computation cost.
\EndFor

\State \textbf{OUTPUT:}
Pareto optimal curve of models at different latency

\State Pick model X and finetune.
\end{algorithmic}
\end{algorithm}

\subsection{Image Classification}

\begin{figure}[t]
  \centering
   \includegraphics[width=0.9\linewidth]{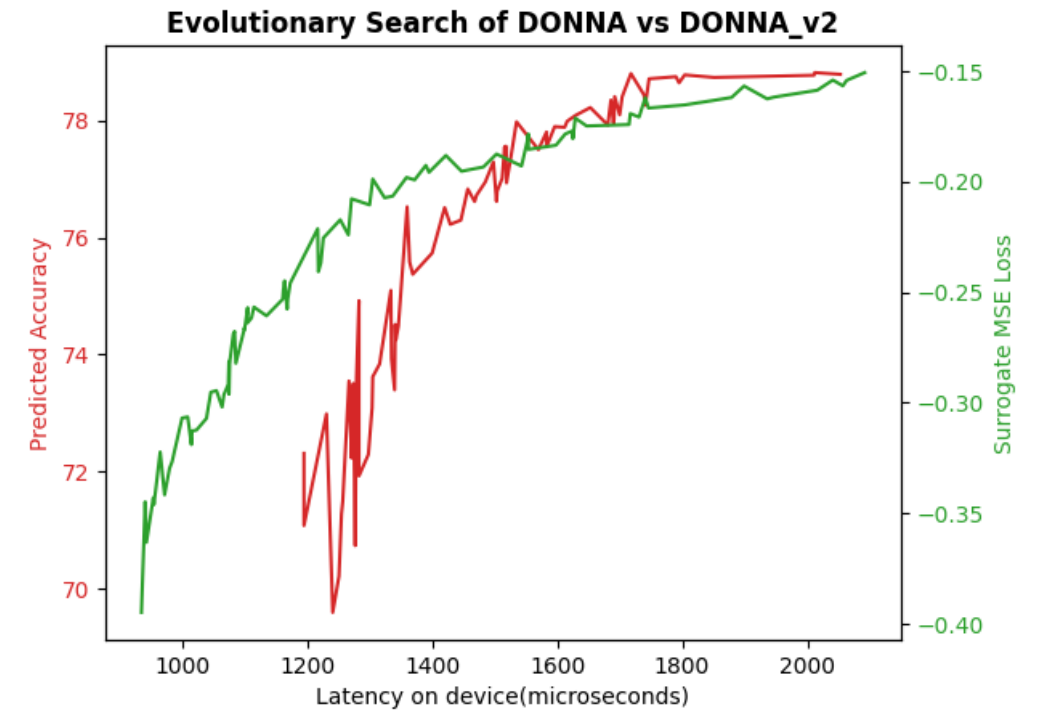}
   \caption{Imagenet Classification Pareto optimal curve of DONNA vs \shortmethodname. The red plot which is the left Y-axis is representative of predicted accuracy vs latency as described in DONNA \cite{donna} and the green plot is the surrogate MSE loss vs latency \textbf{(our proposed method)}. Both the measurements of latency are performed on Samsung Galaxy S10 mobile platform. \shortmethodname search identifies models across a similar latency spread as in the case of DONNA.
   }
   \label{fig:plot}
\end{figure}

We present experiments for DONNA search spaces for ImageNet \cite{imagenet} classification that was earlier discussed in DONNA \cite{donna}. The mothernet chosen here, was Efficientnet-B0 \cite{efficientnet} style architecture with 6 blocks instead of 7. We searched over 5 blocks of the mothernet numbered 1 to 6 using DONNA search space. DONNA search space had a choice out of M=192 options: kernel size $k \in {3,5}$; expansion ratio $expand \in {2,3,4}$; $depth \in {1,2,3,4}$;
$activation \in {ReLU/Swish}$;
$ layer-type \in {grouped, depthwise inverted residual bottleneck}$; and $ channel-scaling \in {0.5, 1.0}$. The search space can be expanded or arbitrarily constrained to known
efficient architectures for a device. Each of these $5 * 192 = 980$ alternative blocks is trained using BKD to complete the Block Library. At this end, we perform the BKD filtering to obtain 768 blocks, thus removing the remaining redundant 212 blocks. After preparing the filtered BKD library, we perform the NSGA-2~\cite{nsga2} algorithm-based search with 100 population size and 50 steps to obtain the Pareto optimal curve. We first show that networks found
by \shortmethodname in the DONNA search space~\cite{donna} outperform the network found by DONNA at similar latency\footnote{Latency numbers could vary by changing the SNPE SDK version. Here we compute the latency of baseline models with a given SDK version and perform NAS with this particular version to observe the compression in latency.}. \shortmethodname achieves similar accuracy at 10X less computational time. The table\ref{tab:Image_classification} shows that the number of epochs for \shortmethodname is significantly lower than DONNA since there is no computation expended for training accuracy predictor. \shortmethodname reduces inference latency as well as model search cost. The model search cost reduction is significant for \shortmethodname since 2500 epochs on ImageNet would cost several GPU hours. Further, in Figure \ref{fig:plot}, we can see that \shortmethodname can identify efficient architectures across a similar latency range as DONNA.
Table\ref{tab:DONNA_vs_perf} captures the comparison of our methodology against the popular NAS methods and it can be observed that our methodology \shortmethodname has lowered the computation cost drastically compared to other methods making it more usable by the research community to find more hardware friendly efficient models. The latency numbers reported in the table \ref{tab:Image_classification} are conducted on Samsung Galaxy S10 mobile platform. Figure \ref{fig:cka_analysis}, describes the efficacy of the block filtering step and compares
models in the Pareto front for DONNA and DONNA v2. The left y-axis is the accuracy predictor stage and the right y-axis is the loss surrogate metric. The figure shows that DONNA v2 search, similar to DONNA, identifies wide range DNN models across the satisfying varying accuracy latency trade-offs. The diversity of models as shown in Figure \ref{fig:cka_analysis} is similar for \shortmethodname and DONNA.

\begin{table*}[h]
\caption{Performances of \shortmethodname compared to other NAS methods}
\label{tab:DONNA_vs_perf}
\begin{center}
\begin{adjustbox}{width=1\textwidth}
\begin{tabular}{|l|c|c|c|c|c|c|}
\hline 
\textbf{Method} & \textbf{Granularity} & \textbf{Macro-Diversity} & \textbf{Search-cost} & \textbf{Cost$/$ Scenario} & \textbf{Cost$/$ Scenario} \\
 \textbf{} & \textbf{}  & \textbf{} &  \textbf{1 scenario [epochs]} &  \textbf{4 scenarios [epochs]} & \textbf{$\infty$ scenarios [epochs]}\\
\hline
DONNA & block-level & variable & 4000 + 10 x 50 & 1500 & 500 \\
\hline
OFA & layer-level & fixed & 1200 + 10 x [25 - 75] & 550 - 1050 & 250 - 750 \\
\hline
NSGANetV2 & layer-level & fixed & 1200 + 10 x [25 - 75] & [550 - 1050] & [250 - 750] \\
\hline
DNA & layer-level & fixed & 770 + 10 x 450 & 4700 & 4500\\
\hline
MNasNet & block-level & variable & 40000 + 10 x 450 & 44500 & 44500\\
\hline
\textbf{\shortmethodname(ours)} & \textbf{block-level} & \textbf{variable} & \textbf{1200} & - & \textbf{500} \\
\hline
\end{tabular}
\end{adjustbox}
\end{center}
\end{table*}

\subsubsection{Performance analysis for classification task}
\begin{figure*}[t]
  \centering
   \includegraphics[width=0.8\linewidth]{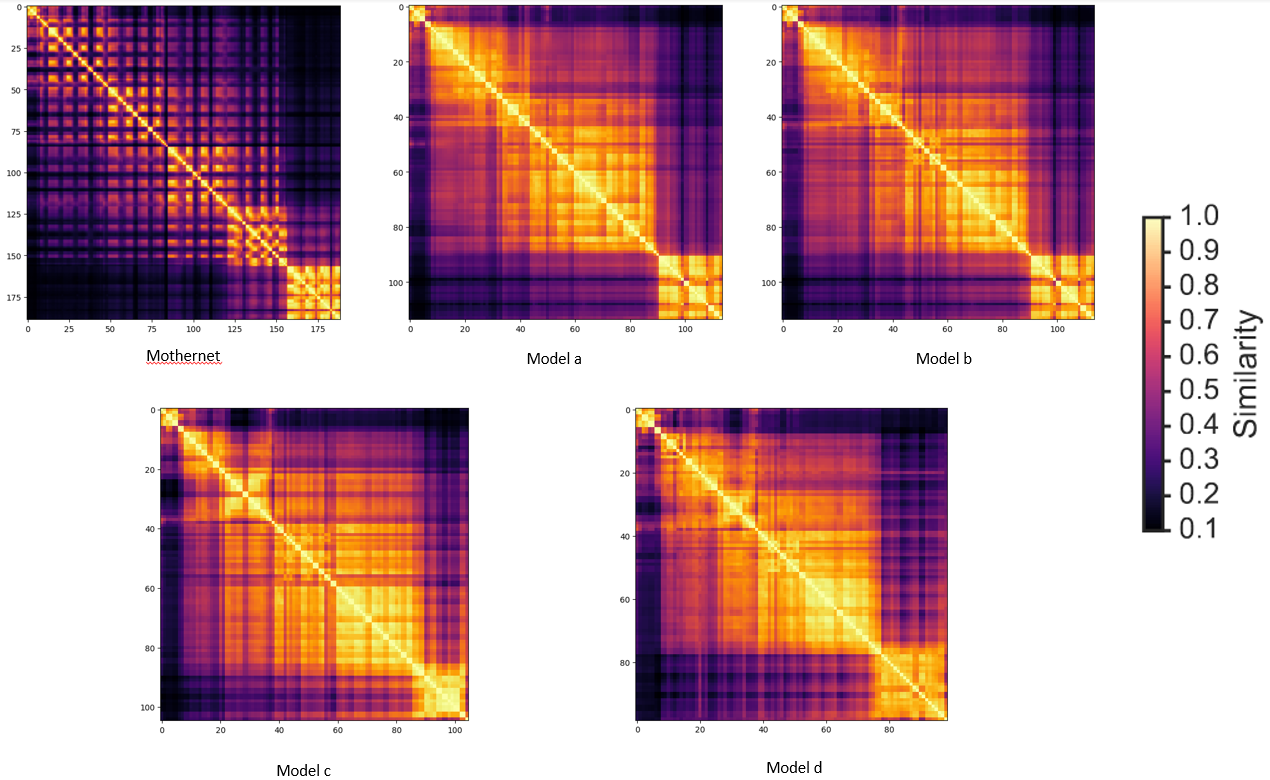}
   \caption{Here, the Mothernet has checkerboard heat map displaying local similarity and model learns different representations across layers. The compressed models (Model a, Model b, Model c and Model d) demonstrate progressively increasing similarity across multiple layers. This suggests that Model a is the best-compressed model in terms of learning distinct representations across layers. This correlates with the performance of the trained model as well as the surrogate loss used in this work.
   }
   
   \label{fig:cka_analysis}
\end{figure*}

Here, we attempt to leverage centralized kernel alignment (CKA), \cite{similarity}, to visualize the \shortmethodname optimized models. Further, we relate interaction between layers of \shortmethodname optimized models using CKA and the surrogate loss. The feature map similarities of CNNs have a block structure. Layers in the same block group (i.e. at the same feature map scale) are more similar than layers in different block groups. \shortmethodname surrogate loss leverages the loss metric of individual blocks forming the network as the performance predictor for the sampled models. CKA analysis of the models from the Pareto optimal curve for the image classification task is shown in figure (\ref{fig:cka_analysis}). In Figure \ref{fig:cka_analysis}, we can observe that the heatmap of layers of the mothernet shows a checkerboard pattern displaying the local block level similarity. The similarity measure for the mothernet is confined to the local blocks.  However, as we start pruning the layers using \shortmethodname, we can observe that for the model (d), a large big yellow box demonstrates similarity in representation across several layers. The fine-tuned accuracy of the model (d) also indicates the performance is saturated. Further, the fine-tuned accuracy of the \shortmethodname optimized models shown in the table \ref{tab:Model_4} correlates with \shortmethodname surrogate loss. The CKA similarity shown in figure(\ref{fig:cka_analysis}) also correlates with \shortmethodname surrogate loss.

\begin{table}[h]
\caption{Performances of \shortmethodname optimized models on Image Classification}
\label{tab:Model_4}
\begin{center}
\begin{adjustbox}{width=0.50\textwidth}
\begin{tabular}{|l|c|c|c|}
\hline
 \textbf{Model} & \textbf{Accuracy} & \textbf{Latency(in ms)} & \textbf{Loss surrogate} \\
\hline
Model a & 78.43  & 1.6 & 0.171 \\
\hline
Model b & 77.8 & 1.47 & 0.188  \\
\hline
Model c & 76.36 & 1.21  & 0.221  \\
\hline
Model d & 74.26 & 1.08  & 0.267  \\
\hline

\end{tabular}
\end{adjustbox}
\end{center}
\end{table}

\begin{table}
\caption{Performances of \shortmethodname optimized models on Image classification}
\label{tab:Image_classification}
\begin{center}
\begin{adjustbox}{width=0.50\textwidth}
\begin{tabular}{|l|c|c|c|c|}
\hline
 \textbf{Model} & \textbf{Accuracy} & \textbf{Latency(in ms)} & \textbf{Cost$/$Scenario} \\
\hline
DONNA & 77.8  & 1.6 & 2500 + 50 + 1e \\
\hline
\textbf{\shortmethodname(ours)} & \textbf{77.8} & \textbf{1.47} & \textbf{50 + 1e}  \\
\hline
EfficientNet-B0 & 77.5 & 2.0  & NA  \\
\hline
\end{tabular}
\end{adjustbox}
\end{center}
\end{table}

\subsection{Object Detection}

Object detection is one of the dense vision tasks, on which extensive neural architecture search is performed. Here, we have identified NAS optimized model EfficientDet-D0 \cite{efficientdet} as the baseline model to further optimize this model in terms of latency and accuracy. The search space identified here has been inspired by the image classification task, as we profiled the object detection model and identified that the majority of the latency of the model resides in the backbone contributing almost 60\% of the end-to-end model. The backbone of the EfficientDet-D0 model is EfficientNet-B0. Hence, our search space for the object detection task includes kernel sizes  $k \in {3,5}$; $expand \in {2,3,4,6}$; $depth \in {1,2,3,4}$;
$ layer-type \in {grouped, depthwise inverted residual bottleneck}$; and $ channel-scaling \in {0.5, 1.0}$. The search space options were expanded for 7 blocks of Efficientnet-b0 \cite{efficientnet} model, making total search complexity to be $128 * 7 = 896$ blocks. Here, we performed the evolutionary search based on NSGA-2 algorithm for 100 population size and 30 steps. The architecture search for object detection performed by us was completely based on MSCOCO \cite{coco} datasets without any imagenet \cite{imagenet} pretraining.  In Table \ref{tab:Object_detection}, we can observe the reduction in computation cost to be around 30\% with improvement in the mAP when compared to the mothernet we started with. This proves that \shortmethodname method can be extended to complex vision tasks like object detection using a loss proxy scoring system to obtain the optimized models from an already compressed NAS searched models like EfficientDet-D0. the latency numbers computed for object detection was performed on Samsung galaxy S10 mobile platform, making this a highly efficient hardware friendly model with better performance as compared to the mothernet we started with.

\begin{table}[h]
\caption{Performances of \shortmethodname optimized models on Object Detection}
\label{tab:Object_detection}
\begin{center}
\begin{adjustbox}{width=0.50\textwidth}
\begin{tabular}{|l|c|c|c|c|}
\hline
 \textbf{Model} & \textbf{mAP} & \textbf{Latency(in ms)} & \textbf{Cost$/$Scenario} \\
\hline
DONNA & 35.1  & 1.98 & 2500 + 310 + 1e  \\
\hline
\textbf{\shortmethodname(ours)} & \textbf{34.8} & \textbf{1.98} & \textbf{310 + 1e}  \\
\hline
EfficientDet-D0 & 33.4 & 2.792 & NA  \\
\hline
\end{tabular}
\end{adjustbox}
\end{center}
\end{table}

\subsection{Super resolution}

For the super-resolution task, we started with a simpler model, Enhanced Deep Residual Networks EDSR \cite{edsr} that has a ResNet-like \cite{resnet} backbone for image super-resolution task. The search space for this model comprised of searching for two blocks based on Resnet Bottleneck style architecture and one head. The search space options for the Resnet style blocks comprised of $depth \in {1,2,3,4,6,8} $ along with input channels and bottleneck channels. The search space options for head were different comprising of kernel sizes and upscaling options. This experiment highlights the use-case that proves that the blocks we chose to optimize need not be similar in architecture to be searched over. We support varying macro-architectural parameters such as layer types, activations and attention mechanisms, as well as micro-architectural parameters such as block repeats, kernel sizes and expansion rates.  Efficient model search for EDSR using DONNA and \shortmethodname resulted in models with comparable performance. However, with \shortmethodname arrived at the efficient EDSR model with 30\%\ reduction in computational cost. REDS \cite{REDS} is a small dataset and the finetuning requires 3125 epochs. To estimate an accuracy predictor for Donna, we subsample and finetune 34 candidate models. \shortmethodname avoids fine-tuning models to estimate the performance of candidate models in the search space. Super-resolution is also one of the dense vision tasks with varying block architectures that was able to converge to optimal models using \shortmethodname search algorithm. 

\begin{table}[h]
\caption{Performances of \shortmethodname optimized models on Super-Resolution}
\label{tab:Super_resolution}
\begin{center}
\begin{adjustbox}{width=0.50\textwidth}
\begin{tabular}{|l|c|c|c|c|}
\hline
 \textbf{Model} & \textbf{PSNR(in dB)} & \textbf{Latency(in ms)} & \textbf{Cost$/$Scenario} \\
\hline
DONNA & 28.36  & 8.68 & 106250 + 3125 + 1e \\
\hline
\textbf{\shortmethodname(ours)} & \textbf{28.44} & \textbf{7.5} & \textbf{3125 + 1e}  \\
\hline
EDSR & 28.6 & 16.7  & NA  \\
\hline
\end{tabular}
\end{adjustbox}
\end{center}
\end{table}

\subsection{Image Denoising}
For image denoising tasks, one of the most popular architectures is the UNet \cite{unet}. To demonstrate the capability of \shortmethodname on image denoising, we chose to optimize a UNet-based multi-stage model NAFNet \cite{nafnet}. NAFnet is one of the state-of-the-art models for image denoising. The evolutionary search over architecture search space with hardware agnostic metrics (Macs count) helped in identifying efficient denoising models with minimal performance degradation. This also demonstrates the efficacy of \shortmethodname for flops-based model search. The optimization strategy could be varied based on use-cases and this is one of the examples proving that \shortmethodname can be performed for a flops-based search strategy as well.
\begin{table}[h]
\caption{Performances of \shortmethodname optimized models on Image Denoising}
\label{tab:denoising_nafnet}
\begin{center}
\begin{adjustbox}{width=0.50\textwidth}
\begin{tabular}{|l|c|c|c|c|}
\hline
 \textbf{Model} & \textbf{GMACs} & \textbf{PSNR} & \textbf{Cost$/$Scenario} \\
\hline
\textbf{\shortmethodname (ours)} & \textbf{40.173} & \textbf{39.9895} & 540+\\
\hline
Stage-1 NAFNet & 63.6 & 40.3045 & NA \\
\hline
\end{tabular}
\end{adjustbox}
\end{center}
\end{table}
Note that NAFNet itself is a lightweight design which added to the difficulty of compressing the model furthermore using NAS. But still, \shortmethodname was able to achieve almost 40\% MAC reduction with only about 0.3 PSNR degradation. It is also one of the complexed vision tasks on which very few NAS methodologies have been applied and proved their efficacy against. 

\begin{table*}[ht!]
\caption{Performances of \shortmethodname optimized models on Multi-task Networks}
\label{tab:MTN_classification}
\begin{center}
\begin{adjustbox}{width=0.95\textwidth}
\begin{tabular}{|l|c|c|c|c|c|}
\hline
\textbf{Model} & \textbf{Detection} & \textbf{Lane segmentation} & \textbf{Driving segmentation area} & \textbf{GMACs} & \textbf{Cost$/$Scenario} \\
\textbf{} & \textbf{mAP} & \textbf{mIOU} & \textbf{mIOU} & \textbf{} & \textbf{} \\
\hline 
\shortmethodname(ours) & 74.7 & 62.5 & 90.7 & 12.4 & 240 + 1e  \\
With backbone alone &  &  &  &  &   \\
\hline
\textbf{\shortmethodname(ours)} & \textbf{75} & \textbf{62.8} & \textbf{91} & \textbf{12.4} & \textbf{240 + 1e}  \\
With head included &  &  &  &  &   \\
\hline
YOLOp & 75.6 & 62.5  & 91.5 & 15.5 & 240  \\
\hline
\end{tabular}
\end{adjustbox}
\end{center}
\end{table*}

\subsection{Multi-Task Network: YOLOP}
For many vision applications, multi-task networks are being deployed and one such widely used model in the autonomous driving industry is YOLOP \cite{yolop}. YOLOP has three tasks: traffic light detection, driving area segmentation, and lane segmentation. The architecture of the model consists of an encoder model which forms the backbone of the network and three heads for each of the tasks. The backbone of YOLOP model comprises of five BottleneckCSP blocks, the object detection head comprises of three BottleneckCSP blocks and the segmentation heads comprise of two BottleneckCSP blocks each. This shows that the computational complexity of the model is spread throughout the model. Here, we explored two approaches to find an efficient compressed model. In the first approach, we compressed only the backbone and in the second approach, the NAS search covered both the backbone and head. In both experiments \shortmethodname helps us come up with networks that are 20-35 \% compressed without significant performance degradation across tasks. The dataset used for the YOLOP network is the BDD100K \cite{bdd100k} dataset. When we compare the compression approaches, as expected, compressing the backbone alone degrades performance across all three tasks when compared with jointly compressing the backbone and the heads. This highlights that when both backbone and heads were searched over, the backbone retained the components needed for accuracy boost and the compression was obtained from the heads as well. This is one of the highly complex model to be searched and it also proves our \shortmethodname can search over segmentation tasks along with multiple head in the tasks as well.

\section{Conclusion}

In this paper, we have explored the efficacy of the surrogate measure and demonstrated a ten-fold reduction in computational complexity for NAS across widely varying learning tasks. It is of great advantage to researchers to be able to perform NAS searches utilizing very few GPU resources. Furthermore, it is important to note that \shortmethodname came up with efficient models while maintaining accuracy across all learning tasks we have explored. Our \shortmethodname was tested extensively across wide range of complex and dense vision tasks and our experimental studies have shown that \shortmethodname has a significant computational advantage for large ImageNet scale training data.  In summary, \shortmethodname provides an efficient NAS approach to building a surrogate performance measure and introduced a novel block filtering approach to improve the quality of models obtained in the evolutionary search step. \shortmethodname has introduced a reliable proxy method that not only makes the NAS faster but can also be applied across wide range of tasks. The limitations of this paper lies in the fact that it is empirically found metric that perform on par or better than the accuracy predictor. In future work, we would like to evaluate the limitations of the metric, if found any.

{\small
\bibliographystyle{ieee_fullname}
\bibliography{egbib}
}

\end{document}